# Программные продукты и системы

/ Программы моделирования и идентификации температурных полей

/ Методы определения весовых коэффициентов для аддитивной фитнес-функции

/ О создании программного комплекса для диагностики расовой принадлежности

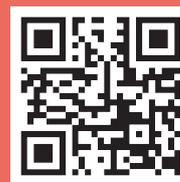

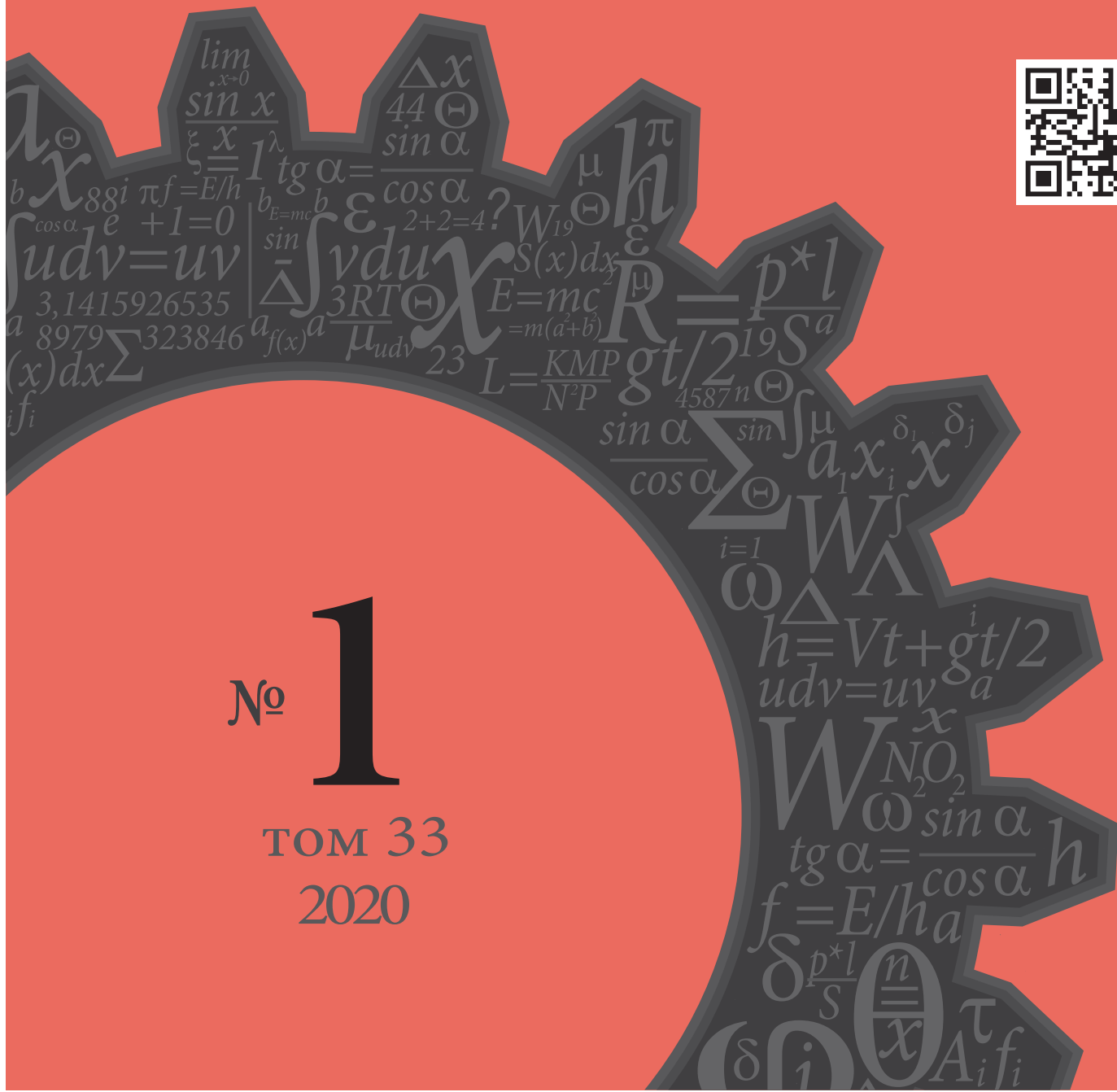

№ 1

том 33

2020

# Программные продукты и системы

# SOFTWARE & SYSTEMS

## *Определение весовых коэффициентов для аддитивной фитнес-функции генетического алгоритма*

*В.К. Иванов* [1], *к.т.н., доцент, начальник управления информационных ресурсов и технологий, mtivk@mail.ru*  
*Д.С. Думина* [1], *магистрант, dumina97@mail.ru*  
*Н.А. Семенов* [1], *д.т.н., профессор кафедры информационных систем, slt1155@mail.ru*

[1] *Тверской государственный технический университет, г. Тверь, 170026, Россия*

Представлено возможное решение задачи выбора способа аналитического определения весовых коэффициентов для аддитивной фитнес-функции генетического алгоритма. Этот алгоритм является основой эволюционного процесса, формирующего в поисковой системе устойчивую и эффективную популяцию запросов для получения высокорелевантных результатов. Приведено формальное описание фитнес-функции алгоритма, которая представляет собой взвешенную сумму трех неоднородных критериев.

Подробно описаны выбранные способы аналитического определения весовых коэффициентов, при этом отмечается невозможность использования методов экспертных оценок. Рассмотрена методика проведения исследований. Описывается исходный набор данных, в том числе диапазоны данных, принятые для вычисления весовых коэффициентов различными способами. Порядок вычислений проиллюстрирован примерами. Результаты исследований, показанные в графической форме, наглядно демонстрируют поведение фитнес-функции при работе генетического алгоритма с использованием различных вариантов весовых коэффициентов.

Анализ результатов позволяет сделать вывод о предпочтительности расчета весовых коэффициентов фитнес-функции данной популяции запросов, выполненного с использованием результатов всех запросов этой популяции. Вывод базируется на наличии последовательных улучшений популяций запросов, характерных для корректной работы генетических алгоритмов, а также на очевидном обнаружении в ходе экспериментов локальных и глобального максимумов фитнес-функции. При использовании других способов расчета весовых коэффициентов подобного не наблюдается. Способ определения весовых коэффициентов для аддитивного критерия оптимальности может повысить качество работы генетического алгоритма для формирования эффективных поисковых запросов. В частности, повышается вероятность быстрого обнаружения локальных экстремумов фитнес-функции, которые на заданной области ее определения могут стать оптимальным решением.

*Ключевые слова: генетический алгоритм, аддитивный критерий, весовой коэффициент, фитнес-функция, хранилище данных, поисковый запрос, релевантность.*

Основная идея технологии генерации поисковых запросов, фильтрации и ранжирования результатов поиска – организация с помощью специального *генетического алгоритма* (ГА) эволюционного *процесса* (ГАП), формирующего в поисковой системе устойчивую и эффективную популяцию запросов для получения высокорелевантных результатов. Специальным образом закодированные запросы последовательно подвергаются генетическим изменениям и выполняются в поисковой системе. Оценивается релевантность промежуточных результатов поиска, вычисляются значения (целевой) фитнес-функции и осуществляется отбор наиболее пригодных запросов. Процесс повторяется до достижения квазиоптимального значения фитнес-функции.

Значение фитнес-функции ГАП определяет качество поисковых запросов и вычисляется для каждого найденного документа в результате выполнения запроса. Это значение зависит от следующих факторов: позиция документа в ранжированном списке результатов запроса, вхождение данного документа в списки результатов других запросов, семантическая близость к адаптивно модифицируемому исходному набору ключевых термов – поисковому паттерну.

В настоящей статье описываются результаты исследования способов аналитического определения веса каждого фактора, влияющего на значение фитнес-функции, и сравнительного анализа применимости каждого способа для оценки динамики изменения значений фитнес-функции ГАП.

### Описание ГАП и его фитнес-функции

В работах [1, 2] отмечается, что в ГАП поисковый паттерн $K$ для документов есть набор термов, относящихся к некоторой предметной области. Каждый поисковый запрос представлен вектором $\bar{q} = (c_1, c_2, ..., c_n, ..., c_m)$, где $c_n = \{k_n, w_n, S_n\}$, $k_n \in K$ – терм; $w_n$ – вес терма; $S_n$ – множество синонимов терма $k_n$; $m$ – количество термов в запросе. Результат выполнения запроса – это набор документов $R$, $|R| = D$. Исходная популяция из $N$ поисковых запросов представлена множеством $Q_0$, где $|Q_0| = N$, $N < |K|/2$, $\bar{q} \in Q_0$. Результатом поискового запроса является множество документов $R$, которое формируется после выполнения $\bar{q}$ в поисковой системе (Bing, Google, БД SQL, данные в структуре XML и т.п.).

Эволюционная операция скрещивания (одно- или двухточечный кроссовер) реализуется обменом термами между запросами, то есть компонентами векторов $\bar{q}$; для репродукции запросов используется генотипный аутбридинг. Адекватная операция мутации – это вероятностная замена синонимом $k'_n \in S_n$ случайно выбранного терма запроса $k_n$. При формировании новой популяции запросов используется элитарный отбор. Условием остановки алгоритма в общем случае считается стабильность популяции.

Значение фитнес-функции, или функции пригодности $\bar{W}$, определяет качество запросов; ГАП ищет максимум $\bar{W}$:

$$\bar{W} = \frac{1}{N}\sum_{j=1}^{N}\bar{w}_j \to \max, \quad (1)$$

где $\bar{w}_j$ – фитнес-функция $j$-го запроса популяции,

$$\bar{w}_j = \frac{1}{R}\sum_{i=1}^{R}w_i(g, p, s), \quad (2)$$

где $w_i$ – фитнес-функция для $i$-го результата $j$-го запроса – результата $r_i$ имеет вид аддитивного критерия оптимальности:

$$w_i = w_g g + w_p p + w_s s. \quad (3)$$

Значение $g$ учитывает ранг для $r_i$, установленный поисковой системой:

$$g = 1 - \frac{g(r_i, R) - g_{\min}}{g_{\max} - g_{\min}}, \quad g(r_i, R) = \sum_{j=1}^{N}pos(r_i)_j^R, \quad (4)$$

где $pos(r_i)_j^R$ – номер позиции $r_i$ в ранжированном списке результатов $j$-го запроса популяции; $g_{\max}, g_{\min}$ – наибольшее и наименьшее значения $g(r_i, R)$ среди всех результатов запросов популяции.

Значение $p$ учитывает универсальность $r_i$, то есть частоту появления $r_i$ в списках результатов других запросов. Оно определяется следующим образом:

$$p = \frac{p(r_i, R) - p_{\min}}{p_{\max} - p_{\min}}, \quad p(r_i, R) = \sum_{j=1}^{N}count(r_i)_j^R, \quad (5)$$

где $count(r_i)_j^R = 1$, если $r_i$ присутствует в списке результатов $j$-го запроса, иначе $count(r_i)_j^R = 0$; $p_{\max}, p_{\min}$ – наибольшее и наименьшее значения $p(r_i, R)$ среди всех результатов запросов популяции.

Значение $s$ определяет семантическую близость $r_i$ и поискового образа $K$. В работе используется косинусная мера близости векторов документов, как это принято в векторной модели пространства документов [3]. Таким образом,

$$s(r_i, K) = \frac{(\bar{v}(r_i)\bar{v}(K))}{\|\bar{v}(r_i)\| \cdot \|\bar{v}(K)\|}, \quad (6)$$

где $\bar{v}(r_i) = \bar{v}(w_1^r, w_2^r, ..., w_n^r, ..., w_{|T|}^r)$ – вектор $i$-го результата запроса, $T$ – количество термов в тексте результата запроса после морфологического анализа (принимаются во внимание только существительные и прилагательные) и лемматизации (в качестве текста результата используются заголовок документа и его краткое описание (сниппет)), $w_n^r = tf_n^r \cdot idf_n^r$ – вес $i$-го терма из текста результата запроса, $tf_n^r$ – частота использования термина в этом тексте, $idf_n^r = \log[(R+1)/R^n]$, $R^n$ – число результатов, текст которых содержит $n$-й терм $i$-го результата; $\bar{v}(K) = \bar{v}(w_1^K, w_2^K, ..., w_m^K, ..., w_{|K|}^K)$ – вектор поискового образа документов $K$, $w_m^K = (1/|K|) \cdot idf_m^K$ – вес $m$-го терма из $K$, $idf_m^K = \log[(R+1)/R^m]$, $R^m$ – число результатов, текст которых содержит $m$-й терм из $K$; $w_g, w_p, w_s$ – весовые коэффициенты для $g, p, s$ соответственно.

### Способы аналитического определения весовых коэффициентов

Как уже отмечалось, задачей являлось исследование способов аналитического определения веса (или значимости) каждого фактора, влияющего на значение фитнес-функции. То есть необходимо определить значения весовых коэффициентов $w_g, w_p$ и $w_s$ для факторных переменных $g, p$ и $s$ в соответствии с (3).

Отметим, что метод взвешенной суммы критериев основан на свертывании всех крите-риев в единственный обобщенный (глобальный, интегральный, агрегированный и т.д.) критерий, представляющий собой сумму критериев, взвешенных коэффициентами их относительной важности, или весами [4]. Метод известен давно, однако до сих пор является довольно распространенным и чаще других используется и активно совершенствуется [5, 6].

Оценка значений весовых коэффициентов с использованием экспертных методов включает следующие основные этапы: определение цели, формирование группы экспертов, разработка сценария и процедур экспертизы, сбор и анализ экспертной информации, анализ результатов экспертизы. Очевидно, что в рассматриваемом случае даже хорошо обоснованные экспертные методы [7] не подходят. Основная причина в отсутствии критериев для совместной сравнительной оценки факторов экспертами. Поэтому разумным представляется принять, что $w_g = w_p = w_s$.

Рассмотрим некоторые способы и приемы, позволяющие по информации о качестве значений факторных переменных определять значения весовых коэффициентов $w_k$ [8, 9].

*Способ 1.* Для каждого частного критерия $F_k(X) > 0$, $k = 1, 2, \ldots, c$, вычисляется коэффициент относительного разброса $\delta_k$, который определяет максимально возможное отклонение по $k$-му частному критерию:

$$\delta_k = \frac{F_k^+ - F_k^-}{F_k^+} = 1 - \frac{F_k^-}{F_k^+}, \quad (7)$$

где $F_k^- = \min_{x \in D} F_k(X)$, $F_k^+ = \min_{x \in D} F_k(X)$. Весовые коэффициенты $w_k$ получают наибольшее значение для тех критериев, относительный разброс которых в области оценок наиболее значительный:

$$w_k = \delta_k \Big/ \sum_{k=1}^{c} \delta_k. \quad (8)$$

*Способ 2.* Пусть все $F_k^- \neq 0$, тогда можно рассмотреть отклонение частного критерия от его наименьшего значения:

$$\beta_k(X) = \frac{F_k(X) - F_k^-}{F_k^-}. \quad (9)$$

Предположим, что важность $k$-го критерия зависит от выполнения неравенства

$$\beta_k(X) \leq \xi_k. \quad (10)$$

Величины $\xi_k$ задаются из условия, что, чем важнее критерий, тем меньшее значение $\xi_k$ выбирается. Геометрическая интерпретация выполнения неравенства (10) будет следующей. Пусть $R_k^*$ – наибольший радиус шара, построенного около точки минимума $x_k^*$ для критерия $F_k(X)$, внутри которого точки $x \in D(x_k^*, \ldots, R_k^*)$ удовлетворяют условию (10). Тогда

$$R_k^* = \max_{x \in D} \left\{ \sum_{k=1}^{c} (x_k - x_k^*)^2 \right\}. \quad (11)$$

Очевидно, что, чем больше радиус шара $R_k^*$, в котором относительное отклонение $k$-го критерия от его минимального значения не превосходит $\xi_k$, тем меньшее значение весового коэффициента $w_k$ надо выбирать:

$$w_k = \frac{1}{R_k^*} \Big/ \sum_{k=1}^{c} \frac{1}{R_k^*}. \quad (12)$$

Также имеются оригинальные разработки методов определения весовых коэффициентов, основанные на эвристических алгоритмах [10].

### Методика проведения исследований

Для вычисления весовых коэффициентов описанными выше способами использовались результаты экспериментов с ГАП, выполненных ранее (https://www.rfbr.ru/rffi/ru/project_search/o_2071601). Исходный набор $K$ был сформирован из терминов предметной области, касающейся управления эволюцией технологических процессов на промышленных предприятиях. Использовались поисковая система Bing и следующий исходный набор значений основных параметров:

– количество запросов в генерируемых популяциях $K = 5$;

– количество ключевых слов в каждом генерируемом запросе $M = 8$;

– максимальное количество результатов поиска, возвращаемых запросом $R_q = 20$, либо популяцией запросов $R_Q = 20$, либо суммарно всеми популяциями $R = 20$;

– вероятность мутации запроса $p_m = 0,1$;

– число проходов алгоритма (или число генерируемых популяций) $N_Q = 200$.

Отметим, что условие остановки алгоритма при проведении экспериментов было отменено для создания условий предотвращения преждевременной сходимости ГАП.

Фрагмент исходных данных в качестве примера представлен в таблице.

Весовые коэффициенты $w_g$, $w_p$ и $w_s$ необходимо вычислить способами 1 и 2. Причем вычисления должны быть произведены для следующих диапазонов исходных данных:

– результаты выполнения каждого запроса популяции, $1 \leq r_i \leq R_q$;

– результаты выполнения запросов каждой популяции, $1 \leq r_i \leq \sum_{q=1}^{K} R_q$;

– результаты выполнения запросов всех популяций, $1 \leq r_i \leq \sum_{N_Q} \sum_{q=1}^{K} R_q$.

**Фрагмент исходных данных**
**Source data chunk**

| № популяции | № особи (запроса) | $g(r_i, R)$ | $g$ | $p(r_i, R)$ | $p$ | $s(r_i, R)$ | $s$ |
|---|---|---|---|---|---|---|---|
| 145 | 559 | 44,00 | 0,43 | 7,75 | 1,00 | 0,05 | 0,42 |
| 145 | 559 | 44,00 | 0,43 | 7,75 | 1,00 | 0,07 | 0,56 |
| 145 | 559 | 26,00 | 0,66 | 2,50 | 0,32 | 0,06 | 0,45 |
| 145 | 559 | 23,00 | 0,70 | 1,75 | 0,23 | 0,06 | 0,46 |
| … | … | … | … | … | … | … | … |
| 145 | 560 | 8,00 | 0,90 | 1,75 | 0,23 | 0,07 | 0,54 |
| 145 | 560 | 4,00 | 0,95 | 1,00 | 0,13 | 0,07 | 0,54 |
| 145 | 560 | 6,00 | 0,92 | 1,00 | 0,13 | 0,07 | 0,55 |
| 145 | 560 | 44,00 | 0,43 | 7,75 | 1,00 | 0,08 | 0,61 |
| … | … | … | … | … | … | … | … |
| 145 | 561 | 21,00 | 0,73 | 1,75 | 0,23 | 0,09 | 0,70 |
| 145 | 561 | 4,00 | 0,95 | 1,00 | 0,13 | 0,07 | 0,51 |
| … | … | … | … | … | … | … | … |

*Порядок вычислений значений весовых коэффициентов по способу 1 (пример).* В качестве исходных примем данные из таблицы. Коэффициенты относительного разброса для частных критериев следующие:

$\delta_g = 1 - (0,43/0,95) = 0,5474$,
$\delta_p = 1 - (0,13/1,00) = 0,8700$,
$\delta_s = 1 - (0,05/0,09) = 0,4444$.

Тогда весовые коэффициенты примут следующие значения:
$w_g = 0,5474/(0,5474 + 0,8700 + 0,4444) = 0,294$,
$w_p = 0,8700/(0,5474 + 0,8700 + 0,4444) = 0,467$,
$w_s = 0,4444/(0,5474 + 0,8700 + 0,4444) = 0,239$.

Отметим, что $w_g + w_p + w_s = 1$.

*Порядок вычислений значений весовых коэффициентов по способу 2 (пример).* В качестве исходных данных также примем исходные данные из таблицы. Пусть $\xi_g = 33$, $\xi_p = 33$, $\xi_s = 34$. Тогда, в соответствии с (10):
$\beta_g = (F(g) - 0,43)/0,43 \leq 0,33$ при $F(g) \leq 0,4719$,
$\beta_p = (F(p) - 0,13)/0,13 \leq 0,33$ при $F(g) \leq 0,3729$,
$\beta_s = (F(s) - 0,42)/0,42 \leq 0,32$ при $F(g) \leq 0,4828$.

Следовательно:
$R_g^* = \max(F(g)) = 0,4719$,
$R_p^* = \max(F(p)) = 0,3729$,
$R_s^* = \max(F(s)) = 0,4828$.

Тогда весовые коэффициенты примут следующие значения:
$w_g = 0,4719/(0,4719 + 0,3729 + 0,4828) = 0,355$,
$w_p = 0,3729/(0,4719 + 0,3729 + 0,4828) = 0,281$,
$w_s = 0,4828/(0,4719 + 0,3729 + 0,4828) = 0,364$.

Отметим, что $w_g + w_p + w_s = 1$.

Вычисленные значения весовых коэффициентов $w_g$, $w_p$ и $w_s = 1$, а также весовые коэффициенты для случая $w_g = w_p = w_s$ должны быть использованы для вычисления фитнес-функции $\overline{W}$. Далее должен быть проведен сравнительный анализ поведения функции $\overline{W}$ при выполнении ГАП с исходным набором $K$.

### Результаты исследований

На рисунке 1 представлены графики зависимости значений фитнес-функции $\overline{W}$ от номеров популяций запросов, порожденных ГАП. Вычисление $\overline{W}$ производилось для документов из диапазона $1 \leq r_i \leq \sum_{N_Q} \sum_{q=1}^{K} R_q$, где $r_i$ – номер $i$-го документа в результатах выполнения запросов всех популяций.

На рисунке 2 изображены графики зависимости значений фитнес-функции $\overline{W}$ от номеров первых 30 популяций запросов, порожденных ГАП. Вычисление $\overline{W}$ производилось для документов из диапазона $1 \leq r_i \leq \sum_{q=1}^{K} R_q$.

На рисунке 3 приведены графики зависимости значений фитнес-функции $\overline{W}$ от номеров первых 10 популяций запросов, порожденных ГАП. Вычисление $\overline{W}$ производилось для документов из диапазона $1 \leq r_i \leq R_q$.

Во всех случаях весовые коэффициенты $w_g$, $w_p$ и $w_s$ принимались равными друг другу, а также вычислялись описанными выше способами 1 и 2. Соответственно, на графиках использованы следующие обозначения фитнес-функции: $W_{equ}$ при $w_g = w_p = w_s$; $W_{dis}$ при вычислении $w_g$, $w_p$ и $w_s$ способом 1; $W_{rad}$ при вычислении $w_g$, $w_p$ и $w_s$ способом 2.

### Обсуждение результатов

Полученные результаты экспериментов позволяют отметить некоторые особенности фитнес-функции ГАП с весовыми коэффициентами, вычисленными различными способами, а также сделать ряд предположений.

1. Как следует из рисунка 1, графики фитнес-функций $W_{equ}$, $W_{dis}$ и $W_{rad}$, значения которых вычислены по результатам выполнения за-

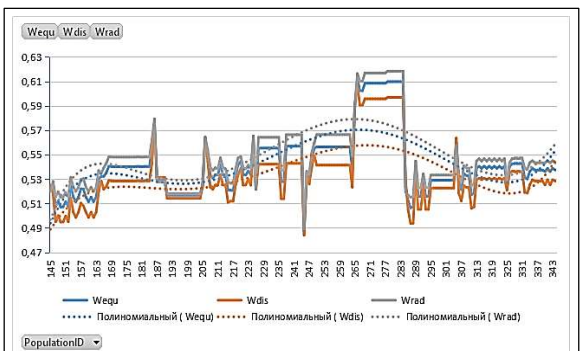

*Рис. 1. Значения фитнес-функции $\overline{W}$, вычисленные по результатам выполнения запросов всех популяций*

*Fig. 1. The values of $\overline{W}$ fitness function calculated by query execution results of all populations*

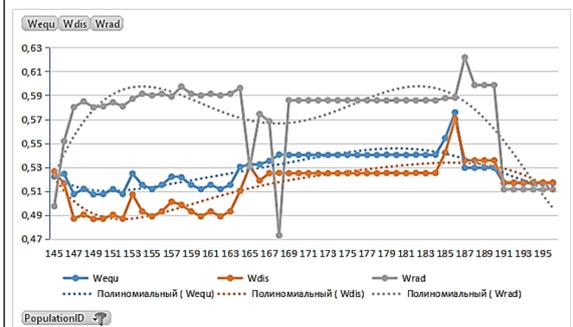

*Рис. 2. Значения фитнес-функции $\overline{W}$, вычисленные по результатам выполнения запросов каждой популяции*

*Fig. 2. The values of $\overline{W}$ fitness function calculated by query execution results of every population*

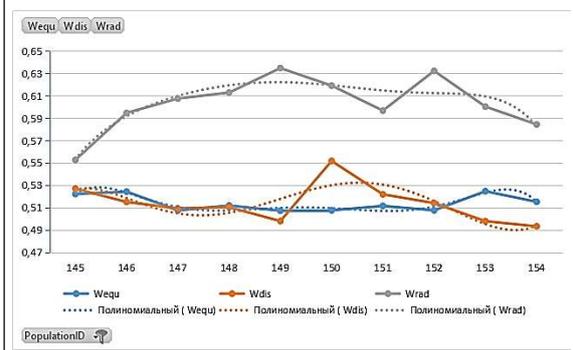

*Рис. 3. Значения фитнес-функции $\overline{W}$, вычисленные по результатам выполнения каждого запроса популяции*

*Fig. 3. The values of fitness function calculated by execution results of every population query*

просов всех популяций, в большой степени похожи. Можно предположить, что $W_{rad} = W_{equ} + \delta_{equ}$ и $W_{rad} = W_{dis} + \delta_{dis}$, причем $\delta_{dis} > \delta_{equ}$ и $R(\overline{W}) \gg R(\delta)$, где $R(\overline{W})$ и $R(\delta)$ – области значений $\overline{W}$ и $\delta$ соответственно. На всех трех графиках отчетливо видны локальный и глобальный максимумы $\overline{W}$, достигаемые практически в одних и тех же точках. В целом при данном диапазоне исходных данных ни один из предложенных способов расчета весовых коэффициентов не дает очевидных преимуществ.

2. При анализе графика функции $W_{rad}$ на рисунке 2 можно увидеть, что при работе ГАП на заданной области определения $W_{rad}$ отчетливо видны два локальных максимума, причем первый из них достигается достаточно быстро (в пределах 10 популяций). Также можно видеть последовательные улучшения популяций запросов, характерные для корректной работы генетических алгоритмов. На графиках функций $W_{equ}$ и $W_{dis}$ подобного не наблюдается: точка первого локального максимума пропущена, точка второго локального максимума совпадает с аналогичной точкой для $W_{rad}$, но сам максимум менее выражен. Вывод – расчет весовых коэффициентов по способу 2 с использованием результатов выполнения запросов каждой популяции представляется более предпочтительным.

3. Графики фитнес-функций $W_{equ}$, $W_{dis}$ и $W_{rad}$ показывают наличие локальных максимумов, найденных ГАП. Однако точки максимумов различны для всех вариантов $\overline{W}$. Из-за небольшого количества шагов выполнения ГАП формулировка каких-либо выводов для данного диапазона исходных данных, используемого при расчете весовых коэффициентов, пока преждевременна.

### Заключение

Результаты экспериментов позволяют сделать вывод об эффективности предложенного подхода. Показано, как метод определения весовых коэффициентов для аддитивного критерия оптимальности – фитнес-функции ГАП – может повысить качество работы генетического алгоритма для формирования эффективных поисковых запросов. В частности, повышается вероятность быстрого обнаружения локальных экстремумов фитнес-функции, которые на заданной области ее определения могут стать оптимальным решением.

Результаты исследования будут использованы при разработке механизма селекции информации об инновационных объектах, основанного на определении семантической релевантности такой информации генерируемым поисковым запросам. Механизм является частью технологии хранилища данных с автоматическим пополнением данными из различных источников.

## Determination of weight coefficients for additive fitness function of genetic algorithm

**V.K. Ivanov** [1], *Ph.D. (Engineering), Associate Professor, Head of Information Resources and Technologies Office, mtivk@mail.ru*  
**D.S. Dumina** [1], *Graduate Student, dumina97@mail.ru*  
**N.A. Semenov** [1], *Dr.Sc. (Engineering), Professor, Information System Department, slt1155@mail.ru*

[1] *Tver State Technical University, Tver, 170026, Russian Federation*

**Abstract.** The paper presents a solution for the problem of choosing a method for analytical determining of weight factors for a genetic algorithm additive fitness function. This algorithm is the basis for an evolutionary process, which forms a stable and effective query population in a search engine to obtain highly relevant results. The paper gives a formal description of an algorithm fitness function, which is a weighted sum of three heterogeneous criteria.

The selected methods for analytical determining of weight factors are described in detail. It is noted that expert assessment methods are impossible to use. The authors present a research methodology using the experimental results from earlier in the discussed project "Data Warehouse Support on the Base Intellectual Web Crawler and Evolutionary Model for Target Information Selection". There is a description of an initial dataset with data ranges for calculating weights. The calculation order is illustrated by examples. The research results in graphical form demonstrate the fitness function behavior during the genetic algorithm operation using various weighting options.

The analysis of the results implies that it is more preferable to calculate fitness function weight factors for this query population then using the results of all population queries. The conclusion is based on the presence of successive improvements in query populations which reflect the correct operation of genetic algorithms, as well as on the obvious detection of local and global maxima in the fitness function during experiments. When using other methods of calculating weighting factors there is no such thing.

Thus, a method for determining weight factors for an additive optimality criterion can improve genetic algorithm quality to generate effective search queries. In particular, the probability of rapid detection of fitness function local extremes is increased and this local extreme can become the optimal solution the function domain.

**Keywords:** genetic algorithm, additive function, weight factor, fitness function, data warehouse, search query, relevance.

***Acknowledgements.*** *The study has been financially supported by the RFBR within the framework of the scientific project no. 18-07-00358 A.*